\documentclass[a4paper, 11pt]{article}
\usepackage[utf8]{inputenc}
\usepackage{graphicx}
\usepackage{amsmath}
\usepackage{caption}
\usepackage{subcaption}
\usepackage{amssymb}
\usepackage[margin=1in]{geometry}
\usepackage{mathtools}
\usepackage{hyperref}
\usepackage{booktabs}
\usepackage{multirow}
\usepackage{authblk}
\usepackage{appendix}
\usepackage{algorithm}
\usepackage{algpseudocode}
\usepackage[backend=biber,sorting=none,style=ieee,natbib]{biblatex}

\hypersetup{linktocpage=true, colorlinks=true, linkcolor=blue, citecolor=blue}
\DeclareUnicodeCharacter{2212}{-}
\title{\textbf{Structure-preserving Planar Simplification for Indoor Environments}}

\author[1]{Bishwash Khanal \footnote{These authors contributed equally to this work.}}
\author[2]{Sanjay Rijal $^*$\footnote{Corresponding authors}}
\author[3]{Manish Awale $^*\dagger$}
\author[4]{Vaghawan Ojha}
\affil[1]{\href{bishwash.khanal@ekbana.info}{\texttt{bishwash.khanal@ekbana.info}}}
\affil[2]{\href{sanjay.rijal@ekbana.info}{\texttt{sanjay.rijal@ekbana.info}}}
\affil[3]{\href{manish.awale@ekbana.info}{\texttt{manish.awale@ekbana.info}}}
\affil[4]{\href{vaghawan.ojha@ekbana.net}{\texttt{vaghawan.ojha@ekbana.net}}}
\affil[ ]{E.K. Solutions Pvt. Ltd., Lalitpur, Nepal}
\date{}

\begin{document}
\maketitle
\providecommand{\keywords}[1]
{
  \small	
  \textbf{\textit{Keywords:}} \textit{#1}
}

\begin{abstract}
  This paper presents a novel approach for structure-preserving planar simplification of indoor scene point clouds for both simulated and real-world environments.
  The scene point cloud initially undergoes preprocessing steps, including noise reduction and Manhattan world alignment, to ensure robustness and coherence in subsequent analyses.   
  We segment each captured scene into structured (walls-ceiling-floor) and non-structured (indoor objects) scenes.
  Leveraging a RANSAC algorithm, we extract primitive planes from the input point cloud, facilitating the segmentation and simplification of the structured scene.
  The best-fitting wall meshes are then generated from the primitives, followed by adjacent mesh merging with the vertex-translation algorithm which preserves the mesh layout.
  To accurately represent ceilings and floors, we employ the mesh clipping algorithm which clips the ceiling and floor meshes with respect to wall normals.
  In the case of indoor scenes, we apply a surface reconstruction technique to enhance the fidelity.
  This paper focuses on the intricate steps of the proposed scene simplification methodology, addressing complex scenarios such as multi-story and slanted walls and ceilings. 
  We also compare qualitative and quantitative performance against popular surface reconstruction, shape approximation, and floorplan generation approaches.
\end{abstract}

\section{Introduction}\label{sec:intro}
With the increasing availability and affordability of high-quality stereo cameras, RGBD sensors, and LiDAR-based cameras,
the reconstruction of raw 3D indoor data has recently emerged as a challenging research problem for generating geometrically accurate and structure-preserving representations.
With the potential to accurately represent indoor environments along with immersive virtual experiences of the real world, 
the demand for simplified models has recently increased in a variety of fields, including architectural design, home decor, real estate marketing, and AR/VR experiences. 
However, modeling complex indoor environments remains a challenging task due to numerous factors such as complex indoor objects, non-planar structured scenes, noises and occlusions in raw data, and so on \cite{rs14194820}.

Our approach begins with a raw point cloud as input, which undergoes axis alignment, scene segmentation, primitives extraction, and simplification to ultimately generate a polygonal mesh.
To simplify the complexity of the scene, we segment the point cloud into structured and non-structured scenes where a structured scene represents the floor, walls, 
and ceiling while non-structured scene refers to indoor objects like tables, desks, beds, and so on.
This allows for separate processing of structured and non-structured scenes: structured scenes are simplified through planar primitive extraction 
and non-structured scenes require surface reconstruction due to intricacies that cannot be represented with planar primitives.
We use RANSAC for extracting plane primitives similar to \cite{rs14174275, li2016manhattan, ksr, nan2017polyfit}. 
However, instead of using candidate faces, hypothesis, and selection strategy \cite{nan2017polyfit, ksr}, we select adjacent planes and enclose them towards their intersection.
This significantly reduces the computational time and the number of faces which we study in detail in section~\ref{sec:results}.
To enhance the geometric representation of the structured scenes, we introduce two novel algorithms: neighboring mesh vertices translation and mesh clipping.
The vertex translation algorithm ensures the enclosedness of neighboring wall meshes by translating their vertices to a common intersection point.
To avoid one-to-all mapping between the wall meshes, we maintain an adjacency list for each wall mesh.
The mesh clipping algorithm, on the other hand, preserves the geometrical structure of ceiling and floor planes by clipping them with respect to adjacent wall meshes.
We assess the performance of our pipeline across simulated, RGBD, and real-world scenes, as described in section ~\ref{sec:results}.
We further provide qualitative and quantitative comparisons with popular shape approximation, surface reconstruction, and floorplan estimation approaches.

In subsequent sections, we provide a literature review in section~\ref{sec:literature}, an overview of our method in section~\ref{sec:overview}, 
detailed explanation of our methodology in section~\ref{sec:method}, experimental results and analysis in section~\ref{sec:results}, 
limitations and future directions in section~\ref{sec:limitations}, and finally, discussion and conclusion in section~\ref{sec:conclusion}.

\section{Related Works} \label{sec:literature}
This section provides a comprehensive review of related research on the simplification of 3D indoor environments, including surface reconstruction, shape approximation, and floorplan generation.

\subsection{Primitive Extraction and Scene Segmentation}
Indoor scene simplification accounts for the inherent complexity of indoor environments, often adopting the Manhattan World (MW) assumption \cite{manhattanworld}. 
This assumption suggests that indoor and urban structures can primarily be described as compositions of 3D orthogonal structures: floors, walls, and ceilings. 
It facilitates the segmentation of large planes within the 3D mesh \cite{app8091529, huang20173dlite, murali2017indoor}. 
Variants such as the Atlanta World \cite{schindler2004atlanta, joo2019globally} are also employed. 
However, these assumptions may not accurately represent spaces with curved or slanted walls, prompting methodologies like point cloud slicing to detect such features \cite{s19173798}.

Simplifying indoor scenes involves two key steps: planar primitive detection and scene segmentation.
Commonly used algorithms for estimating planar primitives from point clouds include RANSAC \cite{RANSAC} and region growing \cite{regiongrowing}.
RANSAC is favored for its efficiency in generating planar primitives, albeit at the cost of accuracy \cite{rs14174275, li2016manhattan, ksr, nan2017polyfit}. 
It can also be used in conjunction with triangulation to recognize and refine planar primitives \cite{8613683}. 
Conversely, region growing offers superior accuracy in terms of primitives refinement \cite{wang2023semantic, s21103493, nikoohemat2020indoor, mura2013robust} at the expense of computational intensity. 
Other notable methods include plane segmentation by clustering point clouds based on saliency analysis \cite{ijgi11040247}, 
Bayesian sampling techniques such as BaySAC \cite{7728063}, and 2D projection-based planar primitive detection \cite{wang2017}.
The 2D projection approach is also utilized to fill missing parts of room layouts \cite{app8091529}. 
Techniques like Hough Transform \cite{paul1962method} and Principal Component Analysis (PCA)-based planar approximation \cite{cai2016surface} offer alternatives for plane detection approaches \cite{wang2018plane}.

Once primitive planes are detected, segmentation of structured and non-structured scenes involves considering 
the angles between the normal plane or its points and the principal coordinate axes \cite{wang2018plane, sanchez2012planar}. 
Geometric and structural constraints are also utilized to identify structural components \cite{rs13193844,fang2021structure}.
Modern deep learning approaches such as PointNet++ \cite{qi2017pointnet++}, and LFCG-Net \cite{rs14194820} leverage semantic information for indoor scene segmentation without the need for primitive extraction. 

\subsection{Surface Reconstruction Approaches}
Several algorithms focus on effectively partitioning space into polygonal faces, convex polyhedra, or computing tetrahedralizations based on intersections of planar primitives. 
\cite{nan2017polyfit} proposes a framework for reconstructing lightweight polygonal faces from point clouds based on a 
hypothesis plane generation followed by selection through linear solvers such as SCIP \cite{BolusaniEtal2024ZR}, GLPK \cite{glpk}, and Gurobi \cite{gurobi}. 
However, as the complexity of the scene increases, it struggles to generate an accurate scene representation due to the large number of candidate faces.
\cite{ksr} employs a shape assembling mechanism utilizing kinetic data structures for space partitioning into convex polyhedra while
Constrained Delaunay Tetrahedralization (CDT) is computed through the intersections of planar primitives \cite{8613683, mura2014reconstructing}.
Despite largely mitigating the computational inefficiencies, it still struggles with complex and real-world scenes,
compromising the scene geometry and generating a large number of polyhedral faces.
 
\cite{fang2021structure} addresses this issue by utilizing a 3D partitioning data structure with a global and local slicing strategy based on a three-level hierarchy:
extracting ceiling and floor primitives, selecting wall planes, and recovering all small planes adjacent to the walls.
This approach generates a compact structure mesh with the reconstruction of indoor and outdoor environments.
While some algorithms specialize in constructing outdoor buildings \cite{chen2022reconstructing, nan2017polyfit, bouzas2020structure}, others focus on indoor properties, leveraging 
methods like merging boxes based on wall classification \cite{murali2017indoor}, Markov Random Field (MRF) formulation \cite{li2016manhattan}, or voxel-based occupancy grid \cite{yang2022polygonal}. 
Additionally, the reconstruction of indoor objects is approached either through traditional surface reconstruction algorithms \cite{fang2021structure, sanchez2012planar} or by replacing objects with existing models using registration \cite{rs14194820}.

\subsection{Shape Approximation Approaches}
Another approach to generating a simplified mesh is to approximate the shape of indoor environments with a small number of faces based on the original shape. With a dense raw 
triangular mesh as input, generally created from surface reconstruction algorithm such as Poisson Surface Reconstruction \cite{kazhdan2013poisson}, a coarser mesh is approximated 
by simplifying the faces until a user-defined criterion is satisfied. The criterion can either be a target number of faces or a geometric error 
between the input and simplified mesh \cite{Garland2023SurfaceSU, 745314}. 
These approaches depend on the edge-collapse operator to iteratively merge the adjacent faces. 
Based on this, adaptive thresholding can be further added to emphasize planar surfaces \cite{garland1999quadric}. 
This quadric error metric (QEM) is enhanced by refining the simplification process to operate planar clusters, 
effectively preserving plane shapes and sharp features while maintaining mesh integrity \cite{8491005}.
Similarly, \cite{rossignac1993vertex} introduces an approach that groups vertices into clusters, and for each cluster, computes a new representative vertex for decimation.
The method proposed in \cite{cohenVSA2004} iteratively computes the regions that best fit the corresponding planar parts and adjusts proxies in each region.
Constrained Delaunay triangulation \cite{huang20173dlite} is also used to triangulate the planar primitives for accurately preserving the boundaries.
However, these shape approximation approaches not being structure-aware may not preserve the original scene geometry.  

\subsection{Floorplan Generation Approaches}
Floorplan approaches focus only on the generation of a 2D layout. Thus, the simplest way to generate a structured 3D model (also popular as a 2.5D model) is to project the 2D layout 
with a predefined camera height. 
These approaches generally rely on either scene segmentation from images \cite{lin2019floorplan, 360dfpe, liu2018floornet, chen2019floorsp} (leveraging geometric segmentation \cite{sun2019horizonnet} and semantic segmentation \cite{lee2017roomnet, hirzer2020smart}) 
or planar primitive patches merging into a planar graph \cite{qi2017pointnet,qi2017pointnet++, he2017mask}.
Approaches like \cite{mura2014automatic} construct a cell complex in the 2D ﬂoor plane followed by individual room partitioning to generate a room polyhedra.

Our method focuses on the piecewise planar primitive reconstruction for structured scenes along with surface reconstruction for non-structured scenes.
Unlike approaches such as \cite{nan2017polyfit, ksr}, we avoid the candidate faces/polyhedra generation and linear solvers that significantly reduce the computational complexity and the number of faces.
Instead of partitioning 3D space, in general, \cite{mura2016piecewise, ksr}, we consider the nature of planar primitives maintaining the local adjacency leveraged by novel vertex translation and mesh clipping
algorithms which further refine the geometry.
Inspired by the hierarchy approach of \cite{fang2021structure}, we use a two-level hierarchy for scene segmentation: extracting structured planes into i) floors and ceiling, and ii) walls.
However, \cite{fang2021structure} fails to incorporate the slanted ceilings and walls, which we address by generating oriented planar meshes with respect to the Z-axis.
In general, our approach reduces the computational costs while still addressing complex scenes like multi-story scenes and scenes with slanted roofs and walls.

\section{Overview}\label{sec:overview}
\begin{figure}[H]
    \centering
    \includegraphics[width=0.55\linewidth, keepaspectratio]{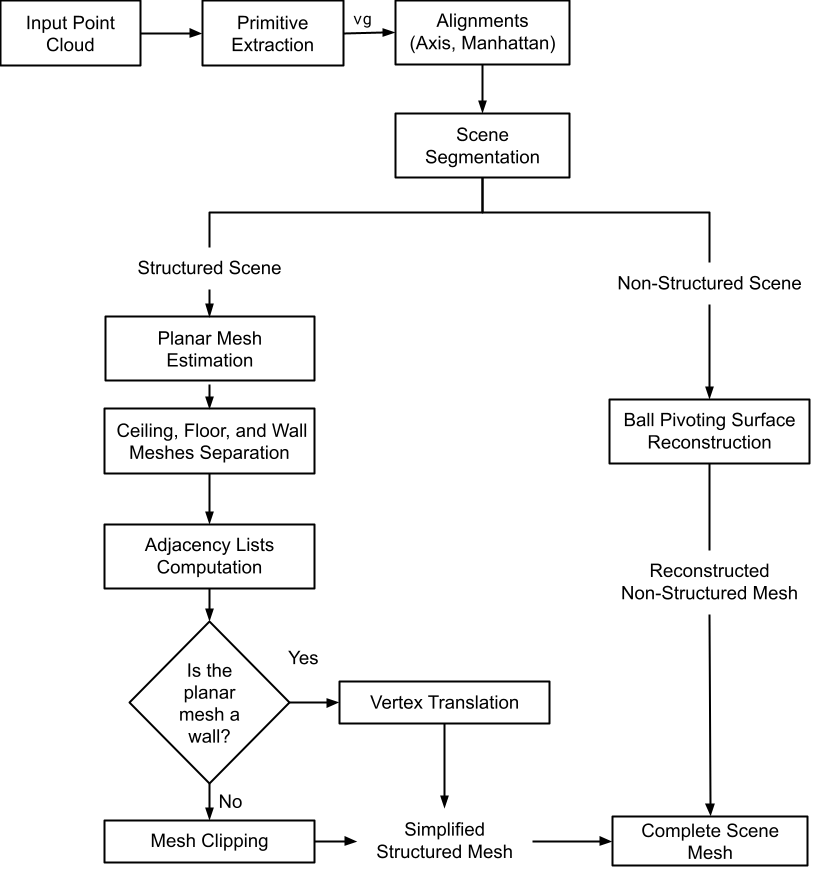}
    \caption{Overall system block diagram of our approach.}
    \label{fig:system-diagram}
\end{figure}

Our pipeline, illustrated in figure \ref{fig:system-diagram}, begins with a raw point cloud as input.
Utilizing the RANSAC algorithm \cite{RANSAC}, we extract planar primitives as vertex groups (\verb|vg|). 
Following the Manhattan World assumption, we align the point cloud along the Z-axis and XY-plane.

The aligned point cloud undergoes segmentation into structured and non-structured primitives, which are then processed separately. 
For structured primitives, due to inaccuracies during data capture and registration, we first generate plane models that best fit the primitive point clouds
followed by the generation of a simplified mesh best representing the plane models.
These simplified meshes may not accurately represent the geometry of the structured scene, 
so we further refine these meshes using vertex translation and mesh clipping algorithms. 

\begin{figure}[ht]
    \centering
    \includegraphics[width=\textwidth, height=\textwidth, keepaspectratio]{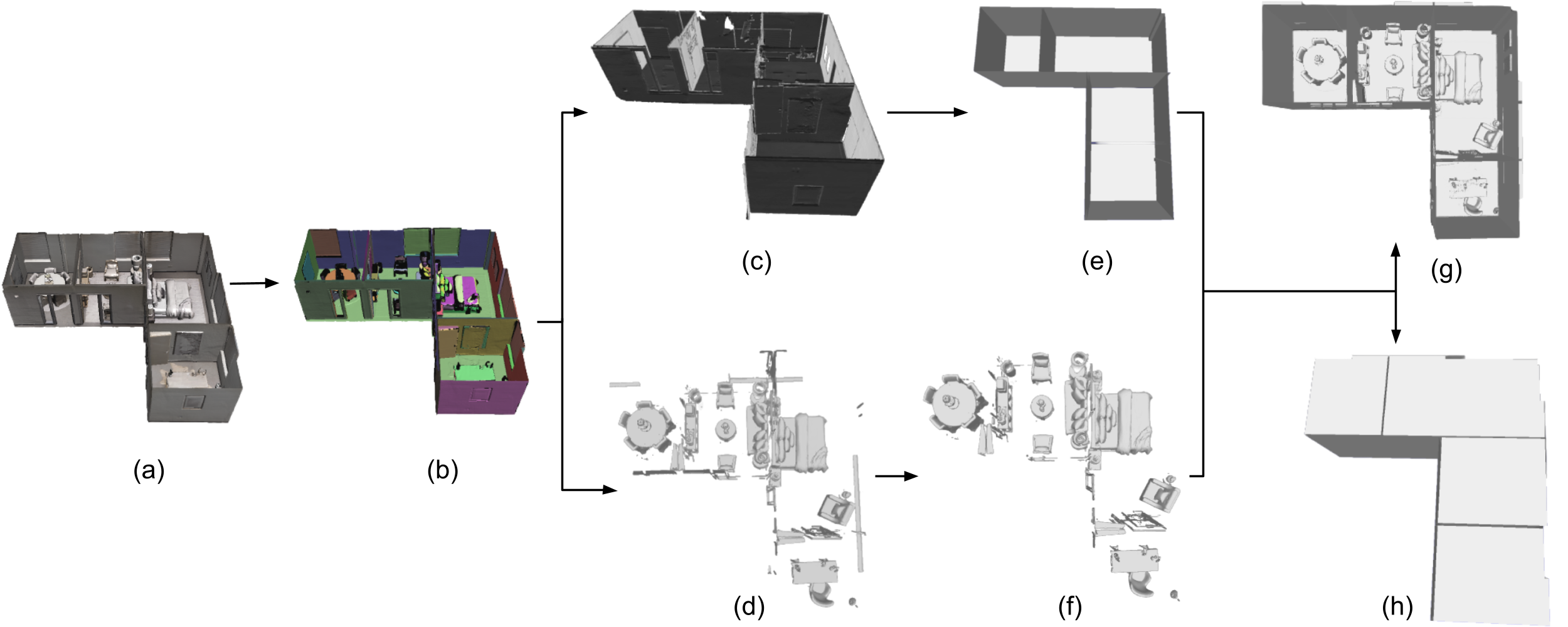}
    \caption{Overview of our approach: (a) input point cloud, (b) planar primitives extraction (as vertex groups), scene segmentation into (c) structured and (d) non-structured scenes,
    (e) generation of simplified structured mesh, (f) surface reconstruction of non-structured scenes, (g) final scene mesh (with (h) its ceiling).}
    \label{fig:overview}
\end{figure}

In cases where partial meshes of a primitive exist due to the absence of a point cloud or error during the data capture, we first establish an adjacency list
for each planar mesh. For each planar wall mesh, we determine the line of intersection between two adjacent meshes and translate the adjacent vertices at the point of intersection.
However, for ceiling and floor meshes, adjacency to all walls complicates accurate representation.
To address this, we implement the mesh clipping algorithm that clips the ceiling and floor meshes iteratively and progressively with the adjacent wall meshes,
thus giving an accurate representation of the scene.
In the case of non-structured scenes, we employ a surface reconstruction technique to enhance the fidelity of the representation. 
These subsequent steps are further illustrated in figure \ref{fig:overview}. 

\section{Methodology} \label{sec:method}
Our method primarily focuses on the steps following the acquisition of input point clouds. The overall process can further be divided into two main parts: scene decomposition, and 
plane estimation and integration.

\subsection{Scene Decomposition}
Scene decomposition mainly involves extracting the planar primitives as vertex groups, alignments, and plane segmentation.
Vertex groups ($\mathcal{P}$), representing plane primitives $\mathcal{P} = \{P_i\}_{i=1,...,n}$, detected by the RANSAC are obtained in \verb|.vg| format which contains 
the points ($x_{ij}, y_{ij}, z_{ij}$) and normals ($\vec{n_i}$) for each vertex group.

\subsubsection{Alignment} \label{axisalign}
For easier scene segmentation and planar mesh estimation, we align the point cloud with respect to the Z-axis (upward direction aligns with the Z-axis) and XY-plane adapted from \cite{li2016manhattan} 
following the Manhattan world assumptions \cite{manhattanworld}. 
This step serves primarily in the planar mesh estimation. 
Although the structured scenes align as per the Manhattan world assumptions, they may contain the primitives unaligned with the primary axes, such as slanted ceilings and walls. 
In such cases, we generate oriented meshes directly from the primitives.
This provides flexibility in the generation of both the axis-aligned and the oriented meshes. 
The steps adopted to align a point cloud along the Z-axis are described in algorithm \ref{alg:zaxis}. 
This algorithm derives a rotation matrix $\mathbf{R}$, which, upon application, rotates the point cloud for alignment with the Z-axis.

\begin{algorithm}[ht]
    \begin{algorithmic}
        \State \textbf{Input:} Planar vertex group list $\mathcal{P} = \{P_i\}_{i=1,...,n}$
        \State \textbf{Output:} Rotation Matrix $\mathbf{R}$ 
        \State Initialize empty lists $U[\;]$ (upward) and $D[\;]$ (downward)
        \For{vertex group $P_{i}$ in $\mathcal{P}$}
            \State Segment $P_{i}$ to get plane parameters $\nu_{i} = [a, b, c, d]$
            \If{$|c| \geq 0.6$}
                \State $[\vec{n_i}]_{z}$ = list of $z$-component of normals of $P_{i}$
                \If{$\texttt{mean}([\vec{n_i}]_{z}) > 0$}
                    \State $U[\;] \mathrel{+}= \nu_{i}$
                \Else
                    \State $D[\;] \mathrel{+}= \nu_{i}$
                \EndIf
            \EndIf
        \EndFor
        \If{$\texttt{len}(U[\;]) > \texttt{len}(D[\;])$}
            \State $\nu_{i} \gets \texttt{max}(\texttt{num\_points}(P_{i}))$
        \Else
            \State $\nu_{i} \gets \texttt{max}(\texttt{num\_points}(P_{i}))$
        \EndIf
        \State Get rotation matrix $\mathbf{R}$ such that vector $\nu_{i} \parallel \hat{k}$
    \end{algorithmic}
    \caption{Alignment Along Z-axis}
    \label{alg:zaxis}
\end{algorithm} 

\begin{figure}[ht]
    \centering
    \includegraphics[width=0.7\linewidth]{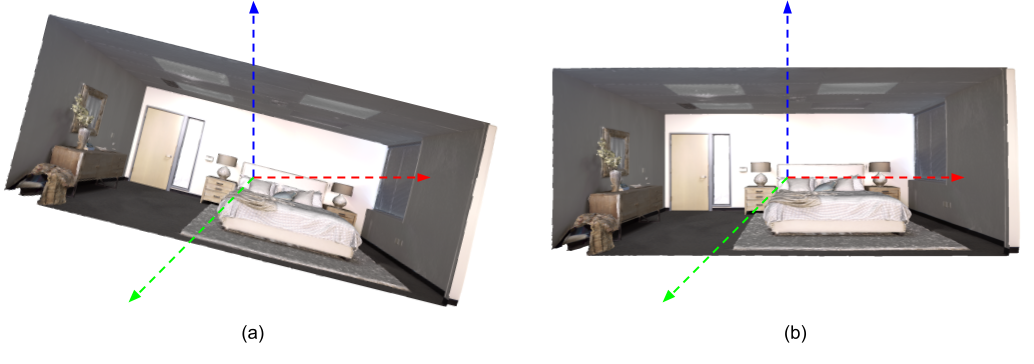}
    \caption{Axis Alignment. (a) Unaligned mesh and (b) Axis-aligned mesh}
    \label{fig:axisalign}
  \end{figure}

\subsubsection{Segmentation}\label{segmentation}
We define segmentation as the differentiation of structured and non-structured scenes.
For this step, we leverage the scene decomposition introduced by \cite{fang2021structure} with a few enhancements as per our requirements
\footnote{We present a detailed ablation study in appendix \ref{sec:ablation_study} encompassing the axis alignment, and thresholds used for vertex translation and mesh clipping algorithms}.
Aligning a scene with the Z-axis enables us to better analyze the distribution of large horizontal planes perpendicular to the Z-axis.

Unlike \cite{fang2021structure} which uses a three-level hierarchy (extracting permanent horizontal structure primitives ($P_{ceiling}$ and $P_{floor}$), 
selecting wall planes ($P_{wall}$), and recovering small structure planes ($P_{small}$)) to categorize structure planes,
we use only the first two levels of hierarchy with some modifications.
Firstly, we consider a primitive $P_i$ as a horizontal plane if angle $\theta (\vec{n_i},\hat{k}) < 20^\circ$, 
where $\hat{k}$ is the normal vector along the Z-axis.
This direct extraction of normals for each primitive obviates the need for the merging and area thresholding techniques described by \cite{fang2021structure}.
From the group of horizontal planes, we consider the largest plane \footnote{largest plane is determined by combination of number of points and volume of the bounding box of the plane} 
along the +Z-axis as $P_{ceiling}$ and along the -Z-axis as $P_{floor}$. 
For scenes with multiple floor and ceiling planes,
\begin{equation}
    \begin{aligned}
        P_i \epsilon P_{ceiling} \mspace{10mu} if \mspace{10mu} h_{P_i} \geq 0.7 h_{\text{max\_ceiling}} \\
        P_i \epsilon P_{floor} \mspace{10mu} if \mspace{10mu} h_{P_i} \geq 0.9 h_{\text{max\_floor}}
    \end{aligned}
\end{equation}
where $h_{P_i}$ is the height of $P_i$, and $h_{\text{max\_ceiling}}$ and $h_{\text{max\_floor}}$ are the heights of the largest ceiling and floor planes respectively. 

For multi-story scenes, $P_i \epsilon P_{ceiling}$ or $P_{floor}$ if it contains at least 10\% of the number of points of the largest horizontal plane. 
Moreover, we consider
\begin{equation}
    \begin{aligned}
        P_i \epsilon P_{walls} \mspace{10mu} if \mspace{10mu} \theta(\vec{n_i},\hat{k}) > 85^\circ \mspace{20mu} and \\
        h_{P_i} > 1.5 m.
    \end{aligned}
\end{equation}

Finally, non-structured point clouds can be extracted by simply subtracting the identified structured point clouds from $\mathcal{P}$, as follows:
\begin{equation}
    P_{non-structured} = \mathcal{P} - P_{structured} = \mathcal{P} - (P_{ceiling} + P_{floor} + P_{walls})
\end{equation}

We also perform statistical treatment on $P_{non-structured}$ for outlier removal before performing surface reconstruction.

\subsection{Plane Estimation and Integration}
Given $P_{structured}$, we estimate the planar meshes, $\mathcal{M} = \{M_{P_i}\}$ for each primitive.
If $\hat{n}_{P_{i}}$ is orthogonal to the coordinate axes, we generate an axis-aligned mesh; otherwise, an oriented mesh is generated.
This gives an exact representation of the orientation of the structured scene.
Given the corresponding adjacency graph $\mathcal{G}_i = \mathcal{(V, E)}$ of a primitive mesh $M_{P_i}$, we process $M_{walls}$ and \{$M_{ceiling}, M_{floor}$\}
\footnote{$M$ represents planar primitive mesh while $\mathcal{M}$ represents output mesh from algorithms: vertex translation and mesh clipping} separately 
to accurately represent the structured scene.
To implement vertex translation and mesh clipping algorithms we mainly leverage \verb|Open3D| \cite{Zhou2018} and \verb|Shapely| \cite{shapely} libraries. 

\subsubsection{Vertex Translation}
\begin{figure}[ht]
    \centering
    \includegraphics[width=0.42\linewidth]{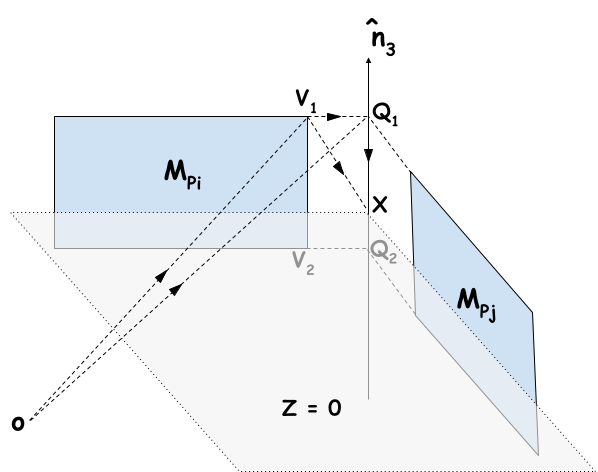}
    \caption{Vertex translation. The vertex $\mathbf{V}_1$ nearest to the line of intersection between $P_i$ and $P_j$ is
    translated to point $\mathbf{Q}_1$. The same process is repeated for vertex $\mathbf{V}_2$.
    $Z=0$ is an arbitrary plane to determine the point of intersection \textbf{X}.}
    \label{fig:vt}
\end{figure}

For $M_{P_i} \; \epsilon \; M_{walls}$, this algorithm first determines the adjacent wall meshes from $\mathcal{G}_i$.
Since the adjacent walls generally intersect with each other, we determine 
the point of intersection, \textbf{X} = $(x, y, z)$ between $M_{P_i}$ plane, its adjacent $M_{P_j}$ plane and an arbitrary plane along $Z=0$ as shown in figure \ref{fig:vt}.
Let $\nu_i$, $\nu_j$ be the plane parameters of $M_{P_i}$, $M_{P_i}$ with normals $\hat{n}_i$, $\hat{n}_j$ respectively.
Then,

\begin{equation}
    A = \begin{pmatrix}
        0 & 0 & 1 \\ 
        \nu_{i0} & \nu_{i1} & \nu_{i2} \\
        \nu_{j0} & \nu_{j1} & \nu_{j2} \\
        \end{pmatrix},
    \;
        B = \begin{pmatrix}
        0 \\ 
        - \nu_{i3} \\
        - \nu_{j3} \\
        \end{pmatrix} 
    \label{eqn:plane_params}
\end{equation}

\begin{equation}
    A\textbf{X} = B \implies \textbf{X} = A^{-1}B
    \label{eqn:intersection_point}
\end{equation}

As shown in figure \ref{fig:vt}, we need to translate the vertex $\mathbf{V}_1$ (nearest to the line of intersection along $\hat{n}_3$) to $\mathbf{Q}_1$ or more specifically 
translate the vector $\vec{OV_1}$ with $\vec{V_1Q_1}$. We first determine the direction of lines where the planes $M_{P_i}$ and $M_{P_j}$ intersect as,
\begin{equation}
    \hat{n}_3 = \hat{n}_i \times \hat{n}_j
    \label{eqn:intersection_direction}
\end{equation}

To avoid the merging of parallel primitives (such as walls on opposite sides of a room), we introduce a parallel primitive threshold, $th_{parallel} = 0.001$.  
$\vec{V_1Q_1}$ is then determined as
\begin{equation}
    \vec{V_1Q_1} = \vec{V_1X} - \vec{Q_1X}
    \label{eqn:line_direction}
\end{equation}
where
\begin{equation}
    \vec{Q_1X} = \textit{projection of } \vec{V_1X} \textit{ on } \hat{n}_3 = \vec{V_1X} \cdot \hat{n}_3 
    \label{eqn:projection}
\end{equation}
 
To avoid the merging of primitives at large separation (such as walls of two rooms separated by a hallway), we introduce a primitive separation threshold, $th_{sep} = 0.5$.
Finally, the extended vector, $\vec{OQ_1}$ is obtained as
\begin{equation}
    \vec{OQ_1} = \vec{OV_1} + \vec{V_1Q_1}
    \label{eqn:vertex_translate}
\end{equation}

\begin{algorithm}[ht]
    \caption{Vertex Translation}
    \begin{algorithmic}
        \State \textbf{Input:} $M_{P_i} \; \epsilon \; M_{walls}$, $\mathcal{G}$
        \State \textbf{Output:} $\mathcal{M}_{walls}$
        \State Initialize variables: $\mathcal{M}_{walls}$[ ], $th_{parallel} = 0.001$, $th_{sep} = 0.5$ 

        \For{$M_{P_i}$ in $M_{walls}$}
            \For{$M_{P_j} \; \epsilon \;  \mathcal{G}_{walls}$}
                \State Compute plane parameters $A$ and $B$ from equation \ref{eqn:plane_params}
                \State Compute the point of intersection, \textbf{X} between $M_{P_i}$ and $M_{P_j}$ from equation \ref{eqn:intersection_point}
                \State Compute the direction of intersection, $\hat{n}_3$ between $M_{P_i}$ and $M_{P_j}$ from equation \ref{eqn:intersection_direction}
                
                \If{$\hat{n}_{3_x} < th_{parallel}$ and $\hat{n}_{3_y} < th_{parallel}$ and $\hat{n}_{3_z} < th_{parallel}$}
                    \State \textbf{continue}
                \EndIf
                \State Compute projection of $M_{P_j}$ on $\hat{n}_3$, $\vec{XQ_1}$ from equation \ref{eqn:projection}
                \State Compute the extending vector $\vec{V_1Q_1}$ from equation \ref{eqn:line_direction}
                \If{$||\vec{V_1Q_1}|| > th_{sep}$}
                    \State $\vec{V_1Q_1}$ = 0
                \EndIf
                \State Extend the vector $\vec{OV_1}$ of $M_{P_i}$ as per equation \ref{eqn:vertex_translate}
            \EndFor
            \State $\mathcal{M}_{walls} \mathrel{+}= M_{P_i}$
        \EndFor
    \end{algorithmic}
\end{algorithm}

\subsubsection{Mesh Clipping}
This algorithm generates geometry-preserving ceiling $\mathcal{M}_{ceiling}$ and floor $\mathcal{M}_{floor}$ meshes 
from their respective primitive planes, $M_{ceiling}$ and $M_{floor}$.
Given the vertex-translated $M_{P_i} \; \epsilon \; \mathcal{M}_{walls}$, this step iteratively clips the $M_{ceiling}$ (and $M_{floor}$)
based on the $\mathcal{G}_{ceiling}$ (and $\mathcal{G}_{floor}$) adjacency.
The approach for clipping $M_{ceiling}$ and $M_{floor}$ is the same, so only the ceiling clipping is explained in this section.

For $M_{P_i} \; \epsilon \; \mathcal{G}_{ceiling}$, the algorithm first determines an axis-aligned bounding box of the \verb|mesh_being_clipped| i.e. ceiling mesh $M_{ceiling}$,
so that only the points enclosed by the box are included for thresholding. 
Using the $\hat{n}_i$ and a vertex position ($x, y, z$) of $M_{P_i}$, the \verb|mesh_being_clipped| is clipped, and the clipper bounding polygons ($poly_1, poly_2$) are determined simultaneously.
We set a threshold, $th_{clip}$ representing the minimum number of points required within the clipper bounding polygons
to include its respective \verb|clipped_mesh_portion| as a part of $M_{ceiling}$. 
Only the \verb|clipped_mesh_portion| satisfying $th_{clip} > 50$ are considered a part of the $\mathcal{M}_{ceiling}$. 
We repeat this process for all $M_{P_i} \; \epsilon \; \mathcal{G}_{ceiling}$ until the $\mathcal{M}_{ceiling}$ is generated.

\begin{algorithm}[ht]
    \caption{Mesh Clipping}
    \begin{algorithmic}
        \State \textbf{Input:} $M_{P_i} \; \epsilon \; \mathcal{M}_{walls}$, $\hat{n}_i$, $M_{ceiling}$, $P_{ceiling}$, $\mathcal{G}_{ceiling}$
        \State \textbf{Output:} $\mathcal{M}_{\textit{ceiling}}$
        \State Initialize variables: $ \mathcal{M}_{ceiling}$[ ], $\texttt{clipped\_mesh\_portion}$[ ], $th_{clip} = 50$, $\texttt{cropped\_pcd}$ 

        \For{$M_{P_i}$ in $\mathcal{G}_{ceiling}$}
            \State $\texttt{clipped\_mesh\_portion} = M_{ceiling}$ 
            
            \For{$\texttt{mesh\_being\_clipped}$ in $\texttt{clipped\_mesh\_portion}$}
                \State Get axis-aligned bounding box of $\texttt{mesh\_being\_clipped}$
                \State $\texttt{cropped\_pcd}$ = crop $P_{ceiling}$ with axis-aligned bounding box 
                \State Clip $\texttt{mesh\_being\_clipped}$ along $\hat{n}_i$ 
                \State $\texttt{clipped\_mesh}$ = clipped parts on positive side of $M_{P_i}.\texttt{vertices}$
                \State $\texttt{clipped\_mesh\_compliment}$ = clipped parts on negative side of $M_{P_i}.\texttt{vertices}$
                \State Pop the first element of $M_{ceiling}$                
                \State Get bounding polygons ($poly_1, poly_2$) of $\texttt{clipped\_mesh}$ and $\texttt{clipped\_mesh\_compliment}$
                \State $poly_1.\texttt{num\_points}$ = $\texttt{cropped\_pcd.points}$ enclosed by $poly_1$
                \State $poly_2.\texttt{num\_points}$ = $\texttt{cropped\_pcd.points}$ enclosed by $poly_2$
                \If{$poly_1.\texttt{num\_points} > th_{clip}$}
                    \State $M_{ceiling} \mathrel{+}= \texttt{clipped\_mesh}$ 
                \EndIf
                \If{$poly_2.\texttt{num\_points} > th_{clip}$}
                    \State $M_{ceiling} \mathrel{+}= \texttt{clipped\_mesh\_compliment}$ 
                \EndIf
            \EndFor
        \EndFor
        \For {$M$ in $M_{ceiling}$}
            \State $\mathcal{M}_{\textit{ceiling}} \mathrel{+}= M$
        \EndFor
    \end{algorithmic}
\end{algorithm}

\begin{figure}[H]
  \centering
  \includegraphics[width=1\linewidth]{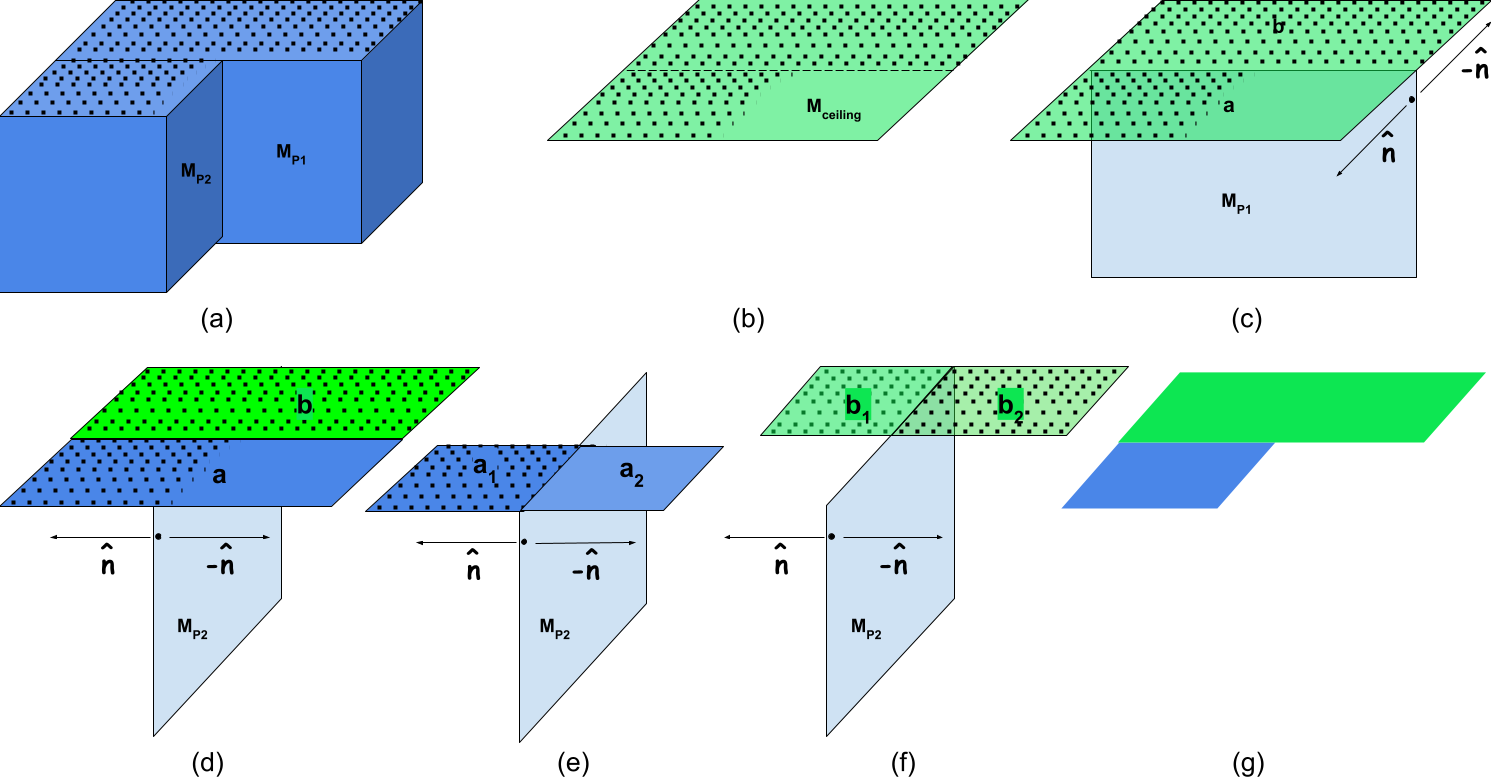}
  \caption{Mesh Clipping. (a) two rooms with intersecting wall meshes, $M_{P1}$ and $M_{P2}$, (b) ceiling plane, $M_{ceiling}$ with points clouds as dots,
  (c) $M_{P1}$ clipping $M_{ceiling}$ into planes \textbf{a} and \textbf{b}, (d)(e) $M_{P2}$ clipping the plane \textbf{a} into planes \textbf{a$_1$} and \textbf{a$_2$},
  (f) $M_{P2}$ clipping the plane \textbf{b} into planes \textbf{b$_1$} and \textbf{b$_2$}, (g) final $\mathcal{M}_{ceiling}$.
  The normal vectors $\pm\hat{n}$ represent the directions of clipping.}
  \label{fig:mc}
\end{figure}

\section{Experimental Results} \label{sec:results}
All the experiments were performed on an Intel i5-6400 CPU @ 2.70GHz paired with 32GB RAM.
The proposed approach largely utilizes Open3D \cite{Zhou2018} for basic geometric computation.

\subsection{Dataset Description} \label{subsec:dataset}
We evaluate our algorithm across a diverse range of datasets, including simulated scenes from the Replica dataset\cite{replica19arxiv}, RGBD scenes 
from HM3D \cite{ramakrishnan2021hm3d} and MP3D \cite{Matterport3D}, as well as custom datasets captured using LiDAR from iPad Pro (M1) and processed with RTAB-Map \cite{Labb__2018}.
Large scene point clouds for custom datasets are generated using Generalized Iterative-Closest Point (ICP) registration \cite{GeneralizedICP}.
In the case of public datasets with scene representations as mesh, we convert the mesh into a point cloud, maintaining the original number of vertices by employing poisson-disk sampling \cite{6143943}. 
We compare our approach against established surface reconstruction, shape approximation, and floorplan generation approaches.

\subsection{Pipeline Evaluation}
\begin{figure}[H]
    \centering
    \includegraphics[width=\textwidth, keepaspectratio]{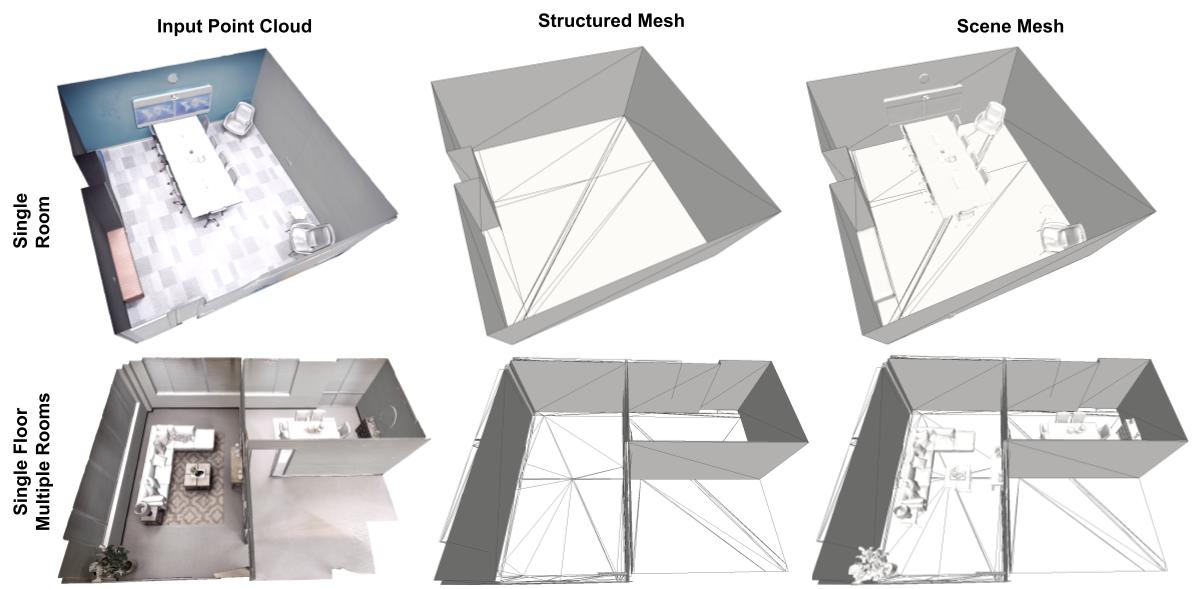}
    \caption{Evaluation on single floor scenes: single room and multiple rooms from the Replica dataset.}
    \label{fig:single_floor}
\end{figure}

\begin{figure}[H]
    \centering
    \includegraphics[width=\textwidth, keepaspectratio]{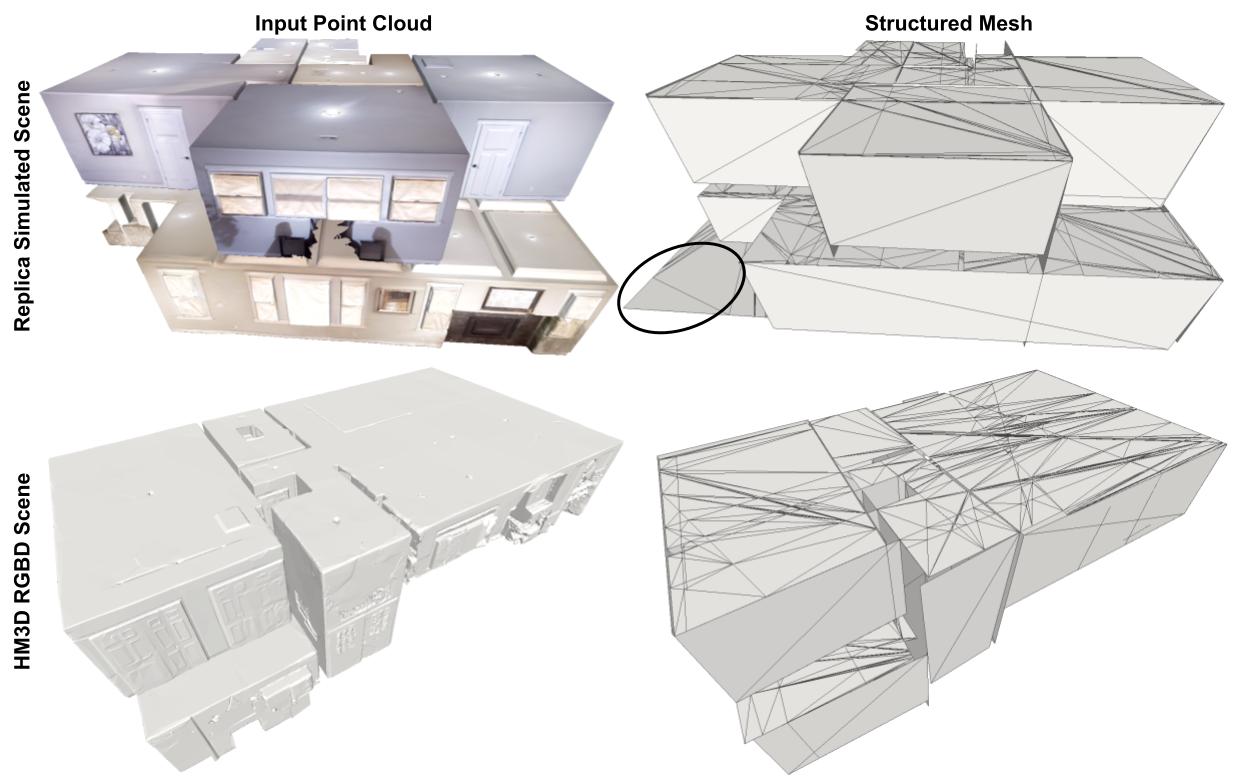}
    \caption{Evaluation on multiple floor scenes from Replica (Simulated) and HM3D (Real-world) datasets.
    The extended mesh within the black region is due to the presence of non-structured objects instead of a wall on the corresponding side point clouds.}
    \label{fig:multiple_floor}
\end{figure}

\begin{figure}[H]
    \centering
    \includegraphics[width=\textwidth, keepaspectratio]{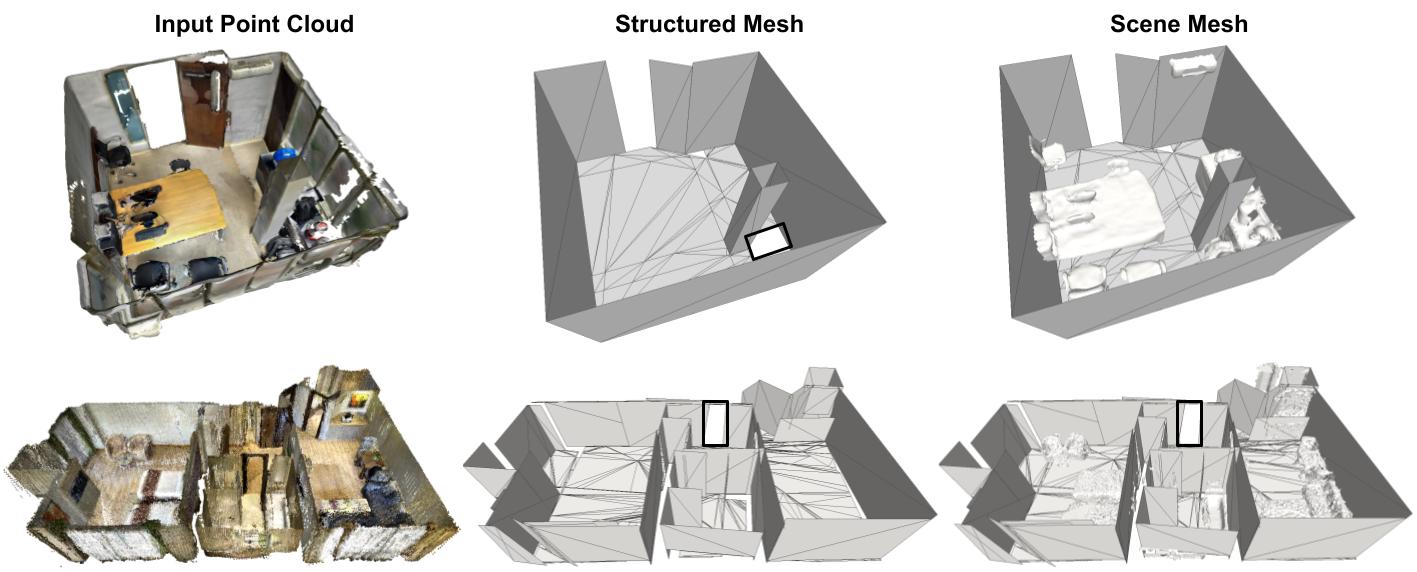}
    \caption{Evaluation on the custom dataset for different scene complexities. The highlighted regions represent sparse or complete lack of points in the input point clouds.
    The black rectangular regions represent the absence of meshes due to sparse or absence of point clouds in raw input.}
    \label{fig:custom_dataset}
\end{figure}

\begin{figure}[H]
    \centering
    \includegraphics[height=0.4\linewidth, keepaspectratio]{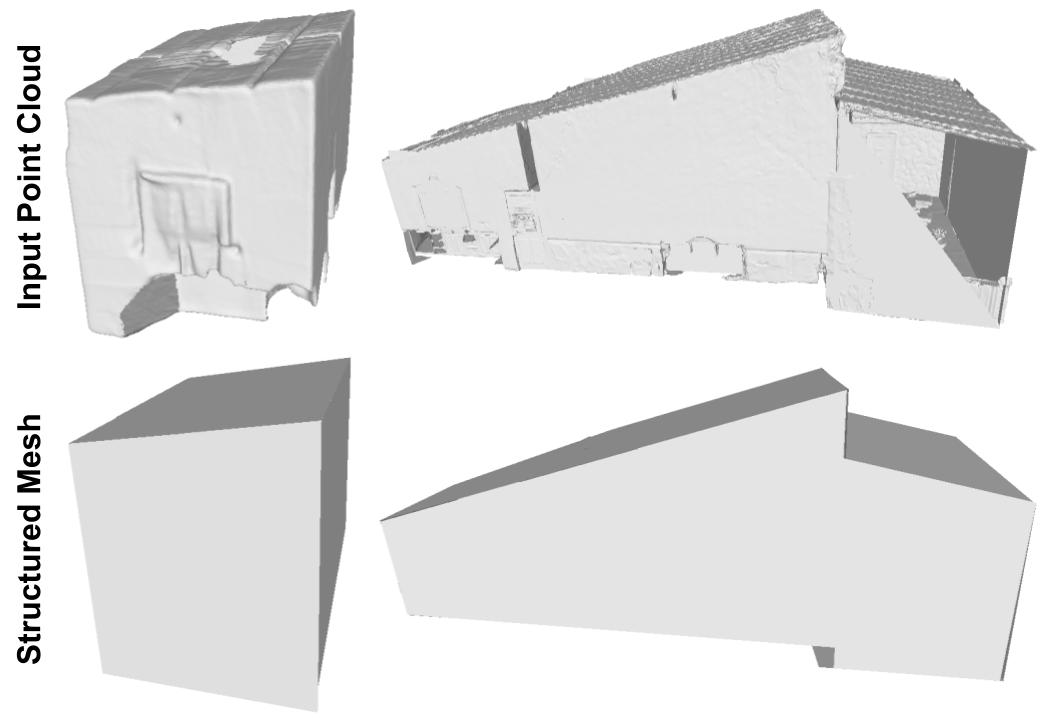}
    \caption{Evaluation on scenes with slanted ceilings and walls.}
    \label{fig:slanted_roof}
\end{figure}

We comprehensively assess the performance of our approach across a variety of scene complexities, encompassing both simulated and custom dataset scenes. 
Ranging from simple single-room environments to intricate multi-story scenes, the evaluation demonstrates the adaptability and robustness of our approach.
Across all evaluated scenes, our approach consistently generates structure-preserving planar simplified scenes as shown in figures \ref{fig:single_floor}, \ref{fig:multiple_floor}, \ref{fig:custom_dataset} and \ref{fig:slanted_roof}. 
For multi-story scenes, we extend our analysis to intermediate floors and ceilings as discussed in Section \ref{segmentation}. 
However, we acknowledge the suboptimal nature of this method for intermediate planes and we plan to improve this in our future enhancements.

We further evaluate the performance of the custom dataset.
Due to the inherent noise from multiple overlapping regions in ICP registration, multiple coplanar planes may be generated for a single planar primitive.
We mitigate this challenge through the appropriate selection of hyperparameters during primitive extraction.
Nonetheless, sparse point clouds can affect the generation of structured planes as shown in figure \ref{fig:custom_dataset}.

While ceilings conventionally exhibit horizontal orthogonality to the floor \cite{s19173798, fang2021structure, mura2013robust}, slanted ceilings pose a unique challenge.
For axis-aligned scenes, our approach also considers horizontal planes tilted at an angle of $\theta \geq 20^{\circ}$ (where $\theta$ is a hyperparameter) along the Z-axis, 
representing the slanted ceilings as planes, as shown in figure \ref{fig:slanted_roof}.
In the case of slanted walls, we reorient the mesh to align with the Z-axis, thereby satisfying the angle criterion.
Overall, our approach preserves the geometry of the scene while also addressing the cases with slanted walls and ceilings.

\subsection{Comparison with Surface Reconstruction Approaches}
\begin{figure}[H]
    \centering
    \begin{subfigure}[b]{\textwidth}
        \includegraphics[width=\textwidth, height=\textwidth, keepaspectratio]{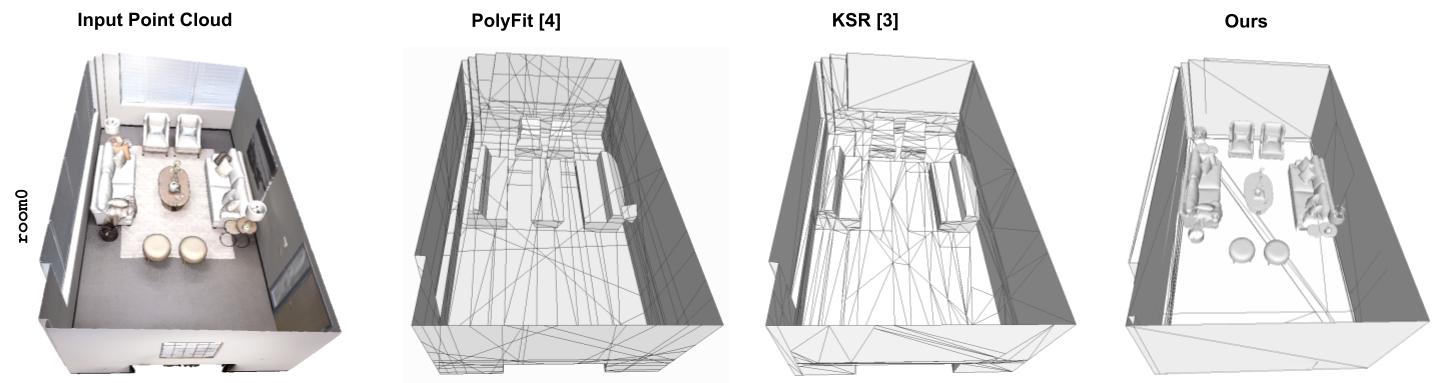}
    \end{subfigure}
    \begin{subfigure}[b]{\textwidth}
        \includegraphics[width=\textwidth, height=\textwidth, keepaspectratio]{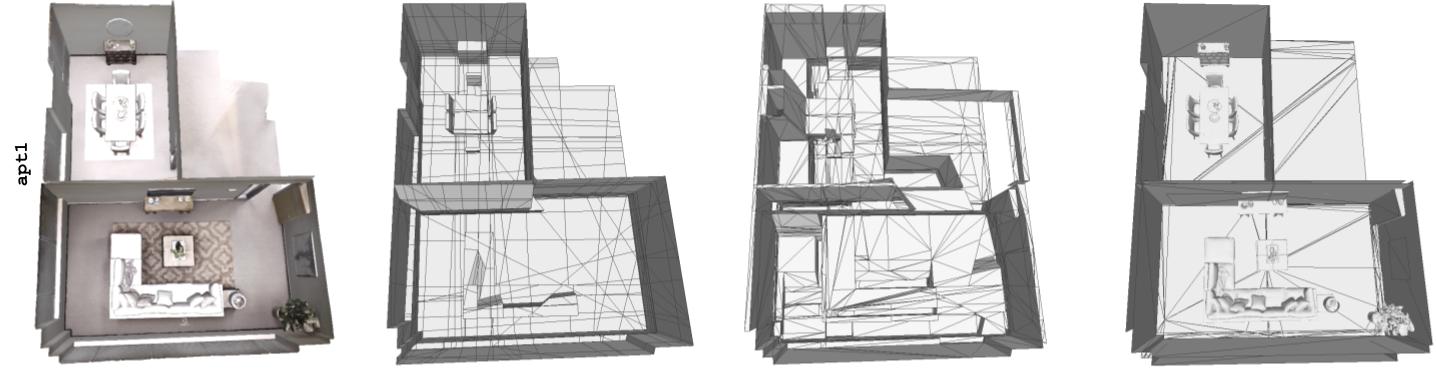}
    \end{subfigure}
    \begin{subfigure}[b]{\textwidth}
        \includegraphics[width=\textwidth, height=\textwidth, keepaspectratio]{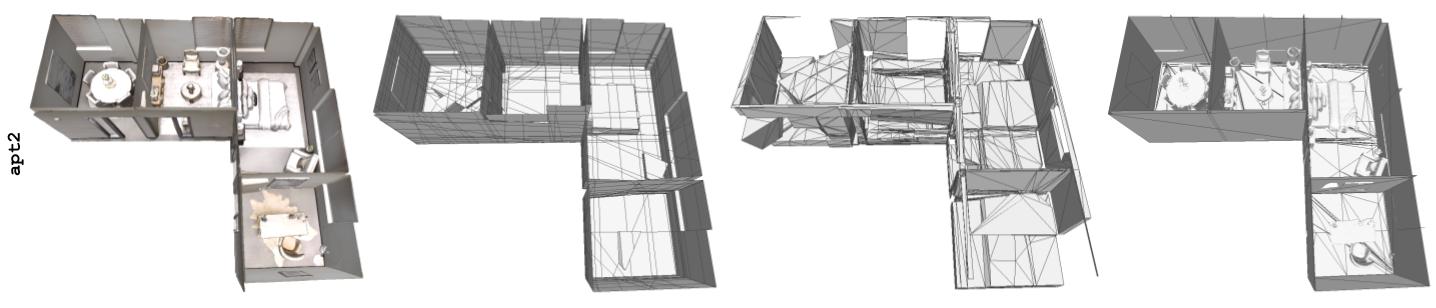}
    \end{subfigure}
    \caption{Comparison with surface reconstruction approaches including both structured and non-structured scenes.}
    \label{fig:replica_comparison}
\end{figure}

We first compare our results with some established surface reconstruction methods such as PolyFit \cite{nan2017polyfit} and Kinetic Shape Reconstruction (KSR) \cite{ksr}
on the Replica dataset scenes.
Our comparison highlights the relative performance of each approach for an equal number of planar primitives, as depicted in figure \ref{fig:replica_comparison}.
Throughout the paper, we calculate Root Mean Square Error (RMSE) using Hausdorff distance \cite{Bryant1970TheCO} as a metric for geometric error between input point cloud and result from different approaches. 

Although KSR can preserve the outer layout of the scene, it fails to accurately reconstruct the inner partitions between the rooms 
such as \verb|apt1| and \verb|apt2| in figure \ref{fig:replica_comparison}, resulting in higher RMSE, as summarized in table \ref{tab:srm_quan}.
Conversely, PolyFit generates a cleaner and more accurate geometry.
However, there exists a trade-off between the scene accuracy and processing time; 
as the scene complexity increases, the computational time significantly increases (in order of days).
This is due to a huge number of candidate faces generated for large complex scenes which is a huge integer programming problem \cite{cgal}.
In contrast, our approach generates geometrically accurate structured scenes in a significantly less amount of time, as tabulated in \ref{tab:srm_quan}.
Moreover, our approach relies on surface reconstruction of indoor objects, thus generating a better representation of indoor objects than the counterparts and consequently a lower RMSE.
\begin{table}[H]
    \centering
    \caption{Comparison of number of faces, $\mathcal{F}$ (for structured scenes) against surface reconstruction approaches. For all scenes, our approach generates a significantly lower number of faces 
    without compromising the scene's accuracy.}
    \begin{tabular}{ccccccccc}
        \toprule
        \multirow{2}{*}{\textbf{Scene}}  & \multirow{2}{*}{\textbf{Initial $\mathcal{F}$}} &\multirow{2}{*}{\textbf{$\mathcal{P}$}} & \multicolumn{3}{c}{Final $\mathcal{F}$}\\ 
        \cline{4-6} &  &  & PolyFit \cite{nan2017polyfit} & KSR \cite{ksr} & Ours  \\ \midrule
        \verb|room0| & 1.9M & 70 & 4263 & 1166 & \textbf{309}  \\ 
        \verb|apt1| & 3.8M & 86 & 6702 & 4864 & \textbf{484}  \\ 
        \verb|apt2| & 4.2M & 101 & 8676 & 5776 & \textbf{820}  \\ 
        \bottomrule
    \end{tabular}
    \label{tab:srm_qual}
\end{table}

\begin{table}[H]
    \centering
    \caption{Quantitive analysis against surface reconstruction approaches using RMSE and computation time (hrs=hours, min=minutes).}
    \begin{tabular}{ccccccccc}
        \toprule
        \multirow{2}{*}{\textbf{Scene}} & \multicolumn{3}{c}{\textbf{RMSE} ($\times 10^{-2}$)} & \multicolumn{3}{c}{\textbf{Computation Time} (approx.)}\\ 
        \cline{2-4} \cline{5-7} &  PolyFit \cite{nan2017polyfit} & KSR \cite{ksr} & Ours &  PolyFit \cite{nan2017polyfit} & KSR \cite{ksr} & Ours \\ \midrule
        \verb|room0| & 0.725  & 1.054  & \textbf{0.334} & 2 hrs & \textbf{3 min} & \textbf{3 min}  \\ 
        \verb|apt1| &  0.688  & 1.267  & \textbf{0.171} & 1 day & \textbf{4 min} & 6 min \\ 
        \verb|apt2| & 0.364  & 0.886 & \textbf{0.031} & 3 days & 6 min & \textbf{5 min} \\ 
        \bottomrule
    \end{tabular}
    \label{tab:srm_quan}
\end{table}

\subsection{Comparison with Shape Approximation Approaches}
We also evaluate our approach against some established shape approximation approaches, including Clustering-based Decimation (CD) \cite{rossignac1993vertex},
Quadratic Edge Collapse-based Decimation (QECD) \cite{garland1999quadric}, and Efficient Plane-based Optimization (plane-opt-rgbd) \cite{8491005}.
We select Replica dataset scenes with complexity spanning from single-room to multi-room environments, including those featuring stairs.
\begin{table}[H]
    \centering
    \caption{Comparison of the number of faces, $\mathcal{F}$ (for structured scenes only). 
    Our approach significantly simplifies the structured scenes compared to other shape approximation approaches (with default parameters).}
    \begin{tabular}{ccccccccc}
        \toprule
        \multirow{2}{*}{\textbf{Scene}} & \multirow{2}{*}{\textbf{Initial $\mathcal{F}$}} &\multicolumn{4}{c}{\textbf{Final} $\mathcal{F}$}\\ 
        \cline{3-6} & &  CD \cite{rossignac1993vertex} & QECD \cite{garland1999quadric} & plane-opt-rgbd \cite{8491005} & Ours \\ \midrule
        \verb|frl_apartment_0| & 825K & 12.2K & 4.7K & 4.7K & \textbf{256} \\ 
        \verb|office_4| & 834K & 26K & 20.7K & 19.3K & \textbf{100} \\  
        \verb|hotel_0| & 696K & 33K & 21.6K  & 9.7K & \textbf{2326} \\  
        \bottomrule
    \end{tabular}
    \label{tab:sam_qual}
\end{table}

\begin{figure}[H]
    \centering
    \includegraphics[width=\textwidth, height=\textwidth, keepaspectratio]{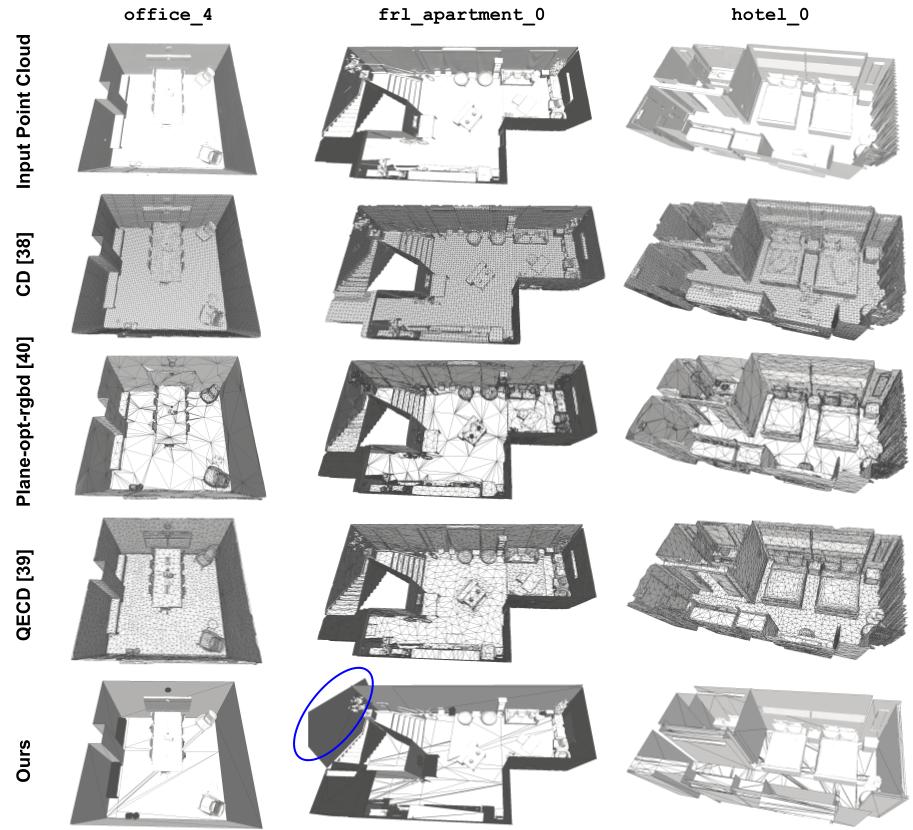}
    \caption{Comparison with shape approximation approaches. The blue region shows the unclipped mesh due to the absence of the ceiling 
    in the original mesh.}
    \label{fig:sam_comparison}
\end{figure}
As shown in figure \ref{fig:sam_comparison}, all approaches generate structure-preserving mesh.
However, focusing on structured scenes, the primary objective of our work, we observe that our approach
significantly reduces the number of faces, as detailed in table \ref{tab:sam_qual}, thus largely simplifying the structured mesh.
Unlike other approaches, our method inherently distinguishes between structured and non-structured scenes.
Thus, for a fair comparison, table \ref{tab:sam_qual} contains a comparison of the number of faces for only the structured scenes.
Notably, in scenes such as $\texttt{frl\_apartment\_0}$ in figure \ref{fig:sam_comparison}, where ceilings are absent, our approach's dependence on clipping ceiling planes 
with respect to neighboring planes may result in unclipped extended planes, as indicated by the blue region.

Additionally, we evaluate RMSE, as shown in table \ref{tab:sam_quan}, including both structured and non-structured scenes. 
RMSE values across all approaches are markedly lower compared to surface reconstruction, owing to the inherent nature of shape approximation approaches, 
which focuses on mesh decimation rather than reconstruction from the original mesh.
In simpler scenes like \verb|office_4|, the lower number of faces in our approach results in lower RMSE. 
However, as scene complexity increases such as \verb|hotel_0|, the number of primitives and faces also increases, leading to elevated RMSE values.

\begin{table}[htb]
    \centering
    \caption{Quantitive analysis against shape approximation approaches. Due to the increase in the number of primitives with increasing scene complexity, RMSE value increases.}
    \begin{tabular}{ccccccccc}
        \toprule
        \multirow{2}{*}{\textbf{Scene}} & \multicolumn{4}{c}{\textbf{RMSE} ($\times 10^{-3}$)}\\ 
        \cline{2-5} &  CD \cite{rossignac1993vertex} & QECD \cite{garland1999quadric} & plane-opt-rgbd \cite{8491005} & Ours  \\ \midrule
        \verb|frl_apartment_0| & 0.665 & 0.334 & 0.763 & \textbf{0.219}  \\ 
        \verb|office_4| & 0.584 & 0.090 & 0.691 & \textbf{0.031} \\
        \verb|hotel_0| & 0.885 & \textbf{0.304} & 0.805 & 0.415 \\
        \bottomrule
    \end{tabular}
    \label{tab:sam_quan}
\end{table}

\subsection{Comparison with Floorplan Generation Approach}
Finally, we evaluate our approach against the widely-used 360-Direct FloorPlan Estimation (DFPE) \cite{360dfpe}.
As the name suggests, floorplan approximation methods inherently generate 2D floorplans, so generally, a pre-defined camera height is assigned to obtain a 3D structured mesh.
Given its superior performance over other popular floorplan estimation approaches like FloorNet \cite{liu2018floornet} and Floor-SP \cite{chen2019floorsp}, we compare our 
results only with 360-DFPE.
We utilize the MP3D dataset \cite{Matterport3D}, which is open-sourced by 360-DFPE, for this comparison.

\begin{figure}[H]
    \centering
    \begin{subfigure}[b]{\textwidth}
        \includegraphics[width=\textwidth, height=\textwidth, keepaspectratio]{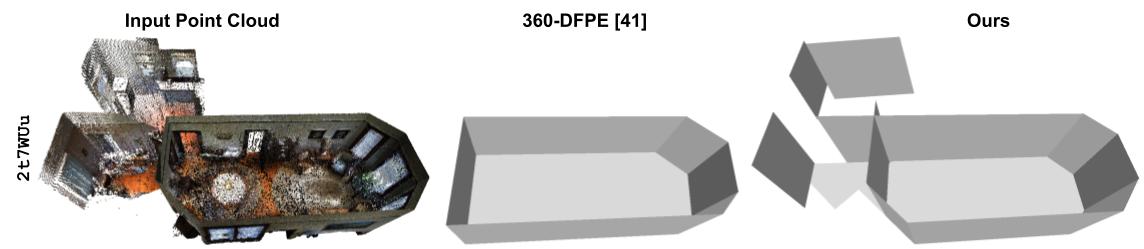}
    \end{subfigure}
    \begin{subfigure}[b]{\textwidth}
        \includegraphics[width=\textwidth, height=\textwidth, keepaspectratio]{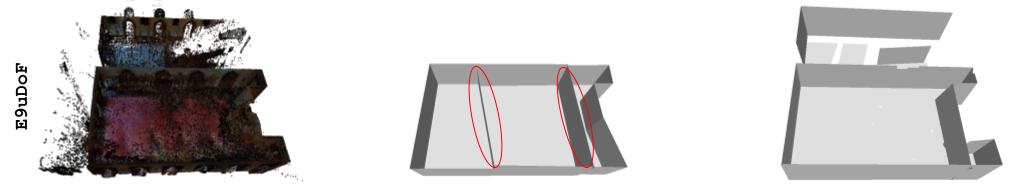}
    \end{subfigure}
    \begin{subfigure}[b]{\textwidth}
        \includegraphics[width=\textwidth, height=\textwidth, keepaspectratio]{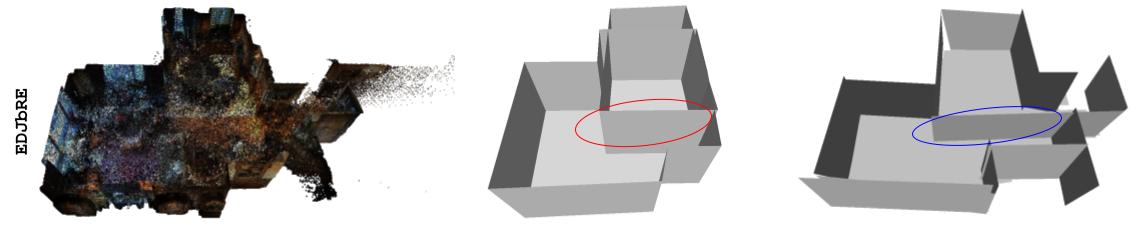}
    \end{subfigure}
    \caption{Comparison with floorplan estimation approach. The wall mesh, highlighted by the blue region, is generated due to a large noise in the input point cloud and
    the mesh, highlighted by the red region, is due to the imperfections (such as occlusions, low-quality images, poor segmentation by HorizonNet, etc.) in wall-ceiling-floor segmentation.}
    \label{fig:dfpe_comparison}
\end{figure}

As illustrated in figure \ref{fig:dfpe_comparison}, 360-DFPE generates a 2D layout only for the enclosed scenes, failing to account for partial rooms. 
Conversely, our approach, leveraging primitive extraction, consistently addresses such scenarios. 
This distinction is further underscored by the RMSE values presented in table \ref{tab:floor_quan}.
Despite minor imperfections observed in our approach due to a significant amount of noise in input point clouds, the overall scene layout is preserved during the wall mesh clipping.

Since our approach relies on the quality of input point clouds for primitive extraction,
in scenes with significant noise like \texttt{EDJbRE} in figure \ref{fig:dfpe_comparison}, our approach may exhibit unwanted wall partitions, as highlighted by the blue region. 
In contrast, 360-DFPE estimates the floorplan from the wall-ceiling-floor segmentation leveraging HorizonNet \cite{sun2019horizonnet}, thus the unwanted wall mesh partitions 
highlighted by the red region in figure \ref{fig:dfpe_comparison}, is due to the imperfections during the segmentation phase and not due to the noises in the input point clouds.  
Comparing the RMSE values in table \ref{tab:floor_quan}, our approach shows significantly lower RMSE.
\begin{table}[H]
    \centering
    \caption{Quantitive analysis against 360-DFPE. Our approach also considers the partial scenes exhibiting a lower RMSE.}
    \begin{tabular}{ccccccccc}
        \toprule
        \multirow{2}{*}{\textbf{Scene}} & \multicolumn{2}{c}{\textbf{RMSE} ($\times 10^{-2}$)}\\ 
        \cline{2-3} &  DFPE \cite{360dfpe} & Ours   \\ \midrule
        \verb|2t7WUu| & 2.8153 & \textbf{0.8258}  \\ 
        \verb|E9uDoF| & 1.1321 & \textbf{0.6089}  \\
        \verb|EDJbRE| & 5.9979 & \textbf{0.5520}  \\ 
        \bottomrule
    \end{tabular}
    \label{tab:floor_quan}
\end{table}

\section{Limitations and Future Works} \label{sec:limitations}
While our approach shows promising performance, certain limitations warrant acknowledgment and present avenues for future research.
The main limitation of our approach lies in the scenes with staircases and curved structured scenes.
As our approach estimates planar meshes solely from planar primitives, it does not account for curved surfaces. 
Additionally, in multi-story scenes, depending on the parameter choices for scene segmentation, our algorithm may struggle to distinguish intermediate ceilings as structured meshes, a limitation that also extends to staircases. 
Addressing these challenges will be a key focus for future improvements, with plans to incorporate capabilities for addressing curved surfaces in subsequent iterations.

Another limitation arises when dealing with a large number of primitives, due to the random nature of RANSAC.
This may result in a single wall mesh being represented by multiple coplanar meshes, thereby introducing noise.
To mitigate this issue, we plan to explore methods for merging smaller meshes close to a wall mesh.
Furthermore, our scene segmentation heavily relies on prior knowledge of the indoor environment, which may lead to mis-segmentation in complex scenes containing features like shelves or wardrobes extending to the ceiling, as structured scenes.
One potential solution is to integrate deep learning-based semantic segmentation with our geometric segmentation approach.

\section{Conclusion} \label{sec:conclusion}
This paper introduces a structured mesh simplification approach based on custom axis alignment, vertex translation, and mesh clipping algorithms. 
Through qualitative and quantitative comparisons with surface reconstruction, shape approximation, and floorplan estimation approaches, we demonstrate the efficacy of our method.
Our approach outperforms the popular surface reconstruction methods like PolyFit and KSR in terms of mesh quality, number of simplified faces, and RMSE.
In comparison to shape approximation techniques, our approach achieves comparable mesh quality while significantly simplifying the structured mesh.
Additionally, our approach offers a more accurate representation of room layouts compared to floorplan approaches, suggesting its potential in floorplan estimation. 
The 3D layouts generated by our pipeline can also be readily projected into 2D floorplans.

Our analysis extends to both simulated (Replica Dataset) and real-world (MP3D, HM3D, and custom dataset captured with iPad Pro (M1)) environments. 
Across both types of data, our approach consistently generates structure-preserving simplified meshes, displaying its versatility in diverse scenarios. 
Furthermore, when combined with texturing algorithms, the simplified meshes derived from our scenes prove useful in real-world applications such as virtual tours, 
architectural design, home decor, and real estate marketing.

\section{Acknowledgements}
The authors would like to extend their sincere gratitude to E.K. Solutions Pvt. Ltd., Nepal for not only providing invaluable support and
resources but also for granting the opportunity to conduct this research. 
We would also like to thank Matterport for their datasets.

\section*{Data Availability Statement}
\href{https://niessner.github.io/Matterport/}{Matterport3D} \cite{Matterport3D} data supporting the results are available from \href{https://github.com/EnriqueSolarte/direct_360_FPE}{direct$\_$360$\_$FPE} GitHub repository
and \href{https://aihabitat.org/datasets/hm3d/}{Habitat Matterport}, and
Replica dataset from \href{https://github.com/facebookresearch/Replica-Dataset}{Replica-Dataset} \cite{replica19arxiv} GitHub repository.
All the custom data generated or appeared in this study are available upon request by contact with the authors.

\section*{Declarations}
All authors certify that they have no affiliations with or involvement in any external organization or entity with any financial or non-financial interest in the subject
matter or materials discussed in this manuscript.

\section*{Author Contributions}
Vertex translation and mesh clipping algorithms--S.R. and M.A.;
Axis alignment algorithm and segmentation of structured and non-structured scenes--B.K.;
Comparisons against surface reconstruction, shape approximation, and floorplan generation approaches--S.R. and B.K.;
Validation, Formal analysis, Software, Investigation M.A., B.K. and S.R.;
Writing original draft preparation--S.R., B.K., and M.A.;
Draft review and editing--S.R., B.K., M. A, and V.O.;
Resources, V.O.;
Supervision, M.A and V.O;
Project administration, M.A, and V.O; 

All authors have read and agreed to the published version of the manuscript.

\newpage
\nocite{*}
\printbibliography

\newpage
\appendix
\renewcommand{\thesection}{\Roman{section}}
\section{Appendix: Ablation Study} \label{sec:ablation_study}

\subsection*{Numbers of Primitives}
\begin{figure}[H]
    \centering
    \includegraphics[width=0.8\textwidth, height=0.8\textwidth, keepaspectratio]{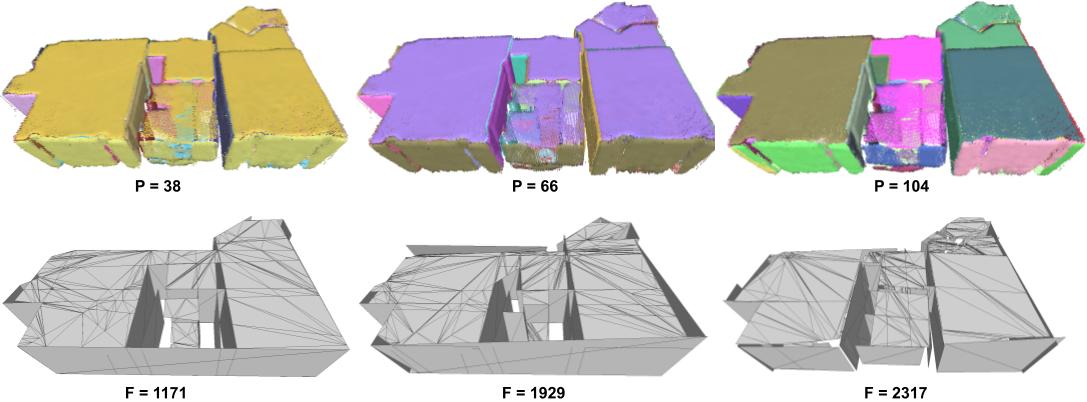}
    \caption{Comparison of the quality of structred mesh generated with varying number of planar primitives.}
    \label{fig:ablation_primitives}
\end{figure}

In this section, we compare the performance of our pipeline on different numbers of primitives.
As shown in figure \ref{fig:ablation_primitives}, with an increasing number of primitives, the number of structured candidate planes also increases.
This increases the outlier meshes thus reducing the quality of the structured mesh.

\subsection*{Axis Alignment}
\begin{figure}[H]
    \centering
    \includegraphics[width=\textwidth, height=\textwidth, keepaspectratio]{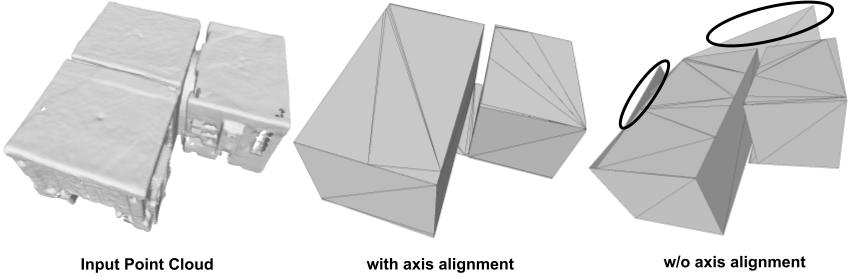}
    \caption{Ablation study on the need for axis-alignment.}
    \label{fig:ablation_axis_alignment}
\end{figure}

We also perform an ablation study focusing on the axis alignment and mesh-clipping algorithm parameters.
Axis alignment is a crucial step in our approach as the selection of ceiling and floor primitives, and generation of simplified meshes (axis-aligned and oriented)
depend on the alignment of the mesh. As shown in figure \ref{fig:ablation_axis_alignment}, the mesh generated without aligning the axes results in some unclipped wall meshes. 

\subsection*{Vertex Translation Algorithm}
The vertex translation algorithm relies on the adjacency lists of individual wall meshes to determine intersecting wall meshes. 
While the intersection between two mesh planes can be determined using the standard equation of planes, it is not always guaranteed that all planes sharing a common intersection point will intersect. 
To address this scenario, we introduce a separation threshold, denoted as $th_{sep}$, which serves as a criterion for accepting intersections between two intersecting meshes and determining the adjacency lists.

\begin{figure}[H]
    \centering
    \includegraphics[width=\textwidth, height=\textwidth, keepaspectratio]{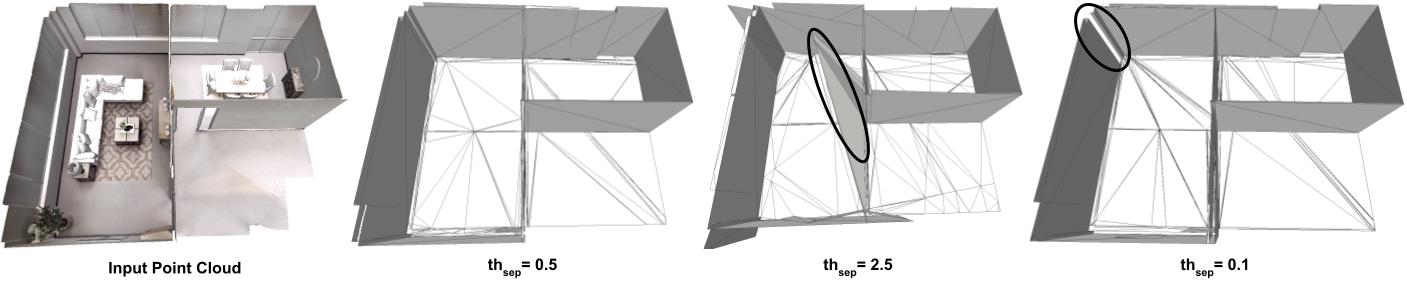}
    \caption{Ablation study on separation threshold parameter, $th_{sep}$ of the vertex translation algorithm.}
    \label{fig:ablation_vertex_translation}
\end{figure}

Figure \ref{fig:ablation_vertex_translation} illustrates the impact of varying $th_{sep}$ on the structured scene mesh. 
In our investigation, we find the optimal $th_{sep}$ to be 0.5. 
Increasing $th_{sep}$ results in the extension of potentially intersecting meshes, causing them to intersect despite the separation present in the input scene
as shown by the black region for $th_{sep}=0.5$ in figure \ref{fig:ablation_vertex_translation}. 
Conversely, decreasing $th_{sep}$ may lead to adjacent meshes, which were expected to intersect, failing to do so, consequently resulting in gaps between the adjacent walls
as highlighted by the black region for $th_{sep}=0.1$ in figure \ref{fig:ablation_vertex_translation}.

\subsection*{Mesh Clipping Algorithm}
\begin{figure}[H]
    \centering
    \includegraphics[width=\textwidth, height=\textwidth, keepaspectratio]{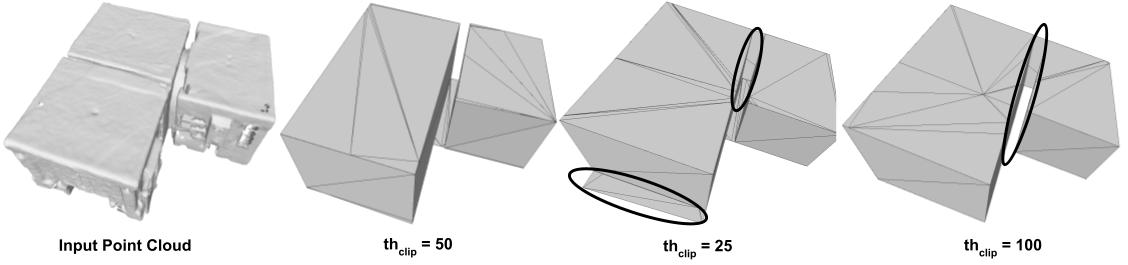}
    \caption{Ablation study on clipping threshold parameter, $th_{clip}$ of the mesh clipping algorithm.}
    \label{fig:ablation_mesh_clipping}
\end{figure}

Ideally, no points should exist in the portion of the mesh to be clipped. However, this is not always true in real-world data; the input point clouds contain noise.
Considering the real-world case, we introduce a clipping threshold parameter $th_{clip}$, which defines the number of point clouds within the portion of the mesh to be clipped.
$th_{clip}$ is a hyper-parameter, thus the choice of the best $th_{clip}$ is a hit-and-trial process. 
In most of our tested scenes, $th_{clip}$ = 50 gave the optimal results.
Figure \ref{fig:ablation_mesh_clipping}, shows the lower value of $th_{clip}$, ($th_{clip}$=25) lefts the unwanted portion of the mesh unclipped while the higher $th_{clip}$ ($th_{clip}$=100) also clips the 
portion of the mesh which is a part of the structured scene. Thus, the choice of $th_{clip}$ significantly affects the quality of the structured mesh. 

\end{document}